\documentclass{acm_proc_article-sp}
\usepackage{enumitem}
\usepackage{hyphenat}
\usepackage{subfigure}
\usepackage{multirow}
\usepackage{booktabs}
\usepackage{acronym}
\usepackage{balance}
\usepackage{array}

\usepackage{fancyhdr}

\newacro{ANN}{Artificial Neural Network}
\newacro{SGD}{Stochastic Gradient Descent}
\newacro{ReSPIR}{Response Surface-based Pareto Iterative Refinement}
\newacro{DSE}{Design Space Exploration}
\newacro{RSM}{Response Surface Modelling}
\newacro{GPU}{Graphics Processing Unit}
\newacro{CPU}{Central Processing Unit}
\newacro{GPGPU}{General-Purpose Computing on \ac{GPU}}
\newacro{MLP}{Multi-Layer Perceptron}
\newacro{CNN}{Convolutional Neural Network}
\newacro{NRE}{Nonrecurring Engineering}
\newacro{ReLU}{Rectified Linear Unit}
\newacro{MPSoC}{Multiprocessor System-on-Chip}
\newacro{FC}{Fully-Connected}
\newacro{ADRS}{Average Distance to Reference Set}
\newacro{PGS-D}{Postgraduate Scholarship - Doctoral}
\newacro{MNIST}[MNIST]{Modified National Institute of Standards and Technology}
\newacro{CIFAR}[CIFAR]{Canadian Institute for Advanced Research}
\newacro{RAM}[RAM]{Random Access Memory}
\newacro{NSERC}[NSERC]{Natural Sciences and Engineering Research Council of Canada}

\global\copyrtyr={\the\year}

\def\sharedaffiliation{
	\end{tabular}
	\begin{tabular}{c}
}

\newcommand*{\refname}{Bibliography}
\begin{document}

\pagestyle{fancy}
\fancyhf{}
\renewcommand{\headrulewidth}{0pt}
\fancyfoot[C]{This is the authors' final version.  The authoritative version is to appear in the ACM Digital Library.}

\setcopyright{acmlicensed}
\conferenceinfo{ICCAD '16,}{November 07 - 10, 2016, Austin, TX, USA}
\isbn{978-1-4503-4466-1/16/11}\acmPrice{\$15.00}
\doi{http://dx.doi.org/10.1145/2966986.2967058}

\title{Neural Networks Designing Neural Networks: Multi-Objective Hyper-Parameter Optimization}

\auwidth=14cm
\numberofauthors{4}
\author{
	\alignauthor{Sean C. Smithson}
	\alignauthor{Guang Yang}
	\alignauthor{Warren J. Gross}
	\alignauthor{Brett H. Meyer}
	\sharedaffiliation
	\affaddr{Department of Electrical and Computer Engineering} \\
	\affaddr{McGill University} \\
	\affaddr{Montreal, Canada} \vspace{1mm} \\
	\email{\normalsize \{sean.smithson, guang.yang3\}@mail.mcgill.ca}, \email{\normalsize \{warren.gross, brett.meyer\}@mcgill.ca}
}

\maketitle

\begin{abstract}
	Artificial neural networks have gone through a recent rise in popularity, achieving state-of-the-art results in various fields, including image classification, speech recognition, and automated control. Both the performance and computational complexity of such models are heavily dependant on the design of characteristic hyper-parameters (e.g., number of hidden layers, nodes per layer, or choice of activation functions), which have traditionally been optimized manually. With machine learning penetrating low-power mobile and embedded areas, the need to optimize not only for performance (accuracy), but also for implementation complexity, becomes paramount. In this work, we present a multi-objective design space exploration method that reduces the number of solution networks trained and evaluated through response surface modelling. Given spaces which can easily exceed $10^{20}$ solutions, manually designing a near-optimal architecture is unlikely as opportunities to reduce network complexity, while maintaining performance, may be overlooked. This problem is exacerbated by the fact that hyper-parameters which perform well on specific datasets may yield sub-par results on others, and must therefore be designed on a per-application basis. In our work, machine learning is leveraged by training an artificial neural network to predict the performance of future candidate networks. The method is evaluated on the \acs{MNIST} and \acs{CIFAR}-10 image datasets, optimizing for both recognition accuracy and computational complexity. Experimental results demonstrate that the proposed method can closely approximate the Pareto-optimal front, while only exploring a small fraction of the design space.
\end{abstract}

\section{Introduction}
\label{sec:introduction}

\ac{ANN} models have become widely adopted as means to implement many machine learning algorithms and represent the state-of-the-art for many image and speech recognition applications \cite{Hinton:DeepLearning}. As the application space for \acp{ANN} evolves beyond workstations and data centres towards low-power mobile and embedded platforms, so too must their design methodologies. Mobile voice recognition systems, such as Apple's \textit{Siri}, currently remain too computationally demanding to execute locally on a handset. Instead, such applications are processed remotely and, depending on network conditions, are subject to variations in performance and delay \cite{SiriNetwork}. \acp{ANN} are also finding application in other emerging areas, such as autonomous vehicle localization and control, where meeting power and cost requirements is paramount \cite{AutonomousVehicleCNN}.

\acp{ANN}, which replace manually engineered computer algorithms, must be trained instead of programmed. This training involves an optimization process where network weights are adjusted with the objective of minimizing the output error. These adjustments often involve a variation of the \ac{SGD} method \cite{Hinton:DeepLearning}. While these training methods have been automated, much of the design process and choice of network hyper-parameters (e.g., number of hidden layers, nodes per layer, or choice of activation functions) has been historically relegated to manual optimization. This relies on human intuition and expert knowledge of the target application in conjunction with extensive trial and error \cite{Domhan:HyperLearningCurves, Young:HyperEvolutionary}. This process is difficult, considering the vast network hyper-parameter space which includes: the number of convolutional or hidden layers, the quantity of nodes in each layer, the type of nonlinear activation functions, and many others which depend on the system in-hand. In addition, the problem with manual hyper-parameter tuning is that there is no guarantee that the process will result in optimal configurations. Not only does the diversity of possible hyper-parameters create extremely large design spaces, but time intensive training phases on comprehensive data sets must also be performed prior to evaluating candidate solutions. This significant computational overhead renders exhaustive searches intractable, and necessitates the use of automated \ac{DSE} tools to intelligently explore the solution space while limiting the number of candidate models that must be trained and evaluated.

\begin{figure*}
	\centering
	\subfigure[Initial configuration with a single hidden layer]{
		\includegraphics[width=0.52\columnwidth]{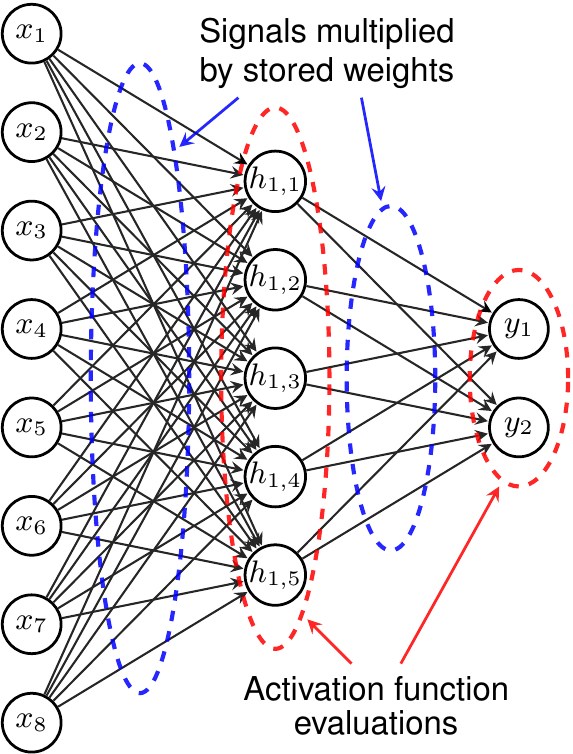}
		\label{fig:MLP1}
	}
	\hspace{0.175\linewidth}
	\subfigure[Reduction in the first stage of multiplications]{
		\includegraphics[width=0.52\columnwidth]{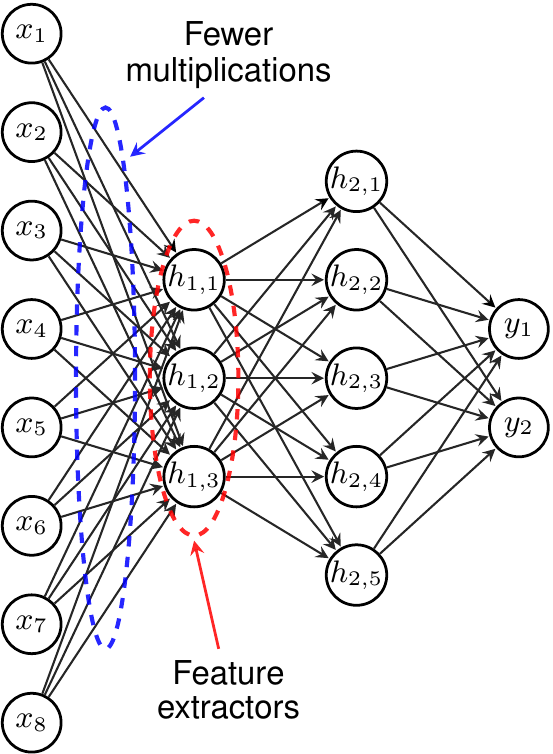}
		\label{fig:MLP2}
	}
	\vspace*{-3mm}
	\caption{Example \acs{MLP} configurations}
	\label{fig:MLP}
\end{figure*}

With the proliferation of machine learning on embedded and mobile devices, \ac{ANN} application designers must now deal with stringent power and cost requirements \cite{Dally:Cost, Hinton:DeepLearning, BigLittle}. These added constraints transform hyper-parameter design into a multi-objective optimization problem where no single optimal solution exists. Instead, the set of points which are not dominated by any other solution forms a Pareto-optimal front. Simply put, this set includes all solutions for which no other is objectively superior in all criteria. Formally, a solution vector $(a_1, b_1)$ (where $a$ and $b$ are two objectives for optimization) is said to dominate another point $(a_2, b_2)$ if $a_1 < a_2$ and $b_1 \le b_2$, or $b_1 < b_2$ and $a_1 \le a_2$; the set of points which are not dominated by any other solution constitutes the Pareto-optimal front \cite{PerformanceIndicesMOO}.

This paper presents an automated \ac{DSE} method that effectively trains a neural network to design other neural networks, optimizing for both performance and implementation cost. This meta-heuristic \ac{ANN} exploits machine learning to predict the performance of candidate solutions (modelling the response surface); the algorithm learns which points to explore and avoids the lengthy computations involved in evaluating solutions which are predicted to be unfit. This leveraging of \ac{RSM} techniques dramatically reduces the proposed algorithm run-time, which can ultimately result in the reduction of product design time, application time-to-market, and overall \ac{NRE} costs.

\subsection{Motivational Example}
\label{sec:motivational_example}

While there are many different \ac{ANN} models, the \ac{MLP} is a well-known form, which rose in popularity with the advent of the back-propagation training algorithm \cite{LinksBetween_MLPs_SVMs}. Characterized by a series of cascaded \textit{hidden} layers, \acp{MLP} have a single feed-forward path from the input to output layers. \ac{MLP} layers are also fully-connected; each node has a connection to each and every node in adjacent layers. When illustrated as a directed graph (shown in Figure~\ref{fig:MLP}), graph connections are representative of signals being multiplied by corresponding weights, and graph nodes the summation of all inputs followed by non-linear activation functions. The evaluation of such elements, commonly \acp{ReLU} or sigmoid functions, is comparatively simple. Therefore, multiply-accumulate operations and memory accesses remain the dominant tasks in terms of cost \cite{Dally:Cost, Jarrett:BestMultiStageArchitecture}.

While structurally simple, determining an optimal \ac{MLP} configuration is a difficult task due to the large number of parameters inherent to even the simplest designs. For example, given the simple \ac{MLP} shown in Figure~\ref{fig:MLP1}, with $i$ inputs, a single hidden layer with $j$ nodes, and $k$ nodes in the output layer, $i \times j + j \times k$ operations must be evaluated at each inference pass and an equal number of weights must be accessed in memory. Starting from this initial configuration, a designer may then choose to include a second hidden layer, as illustrated in Figure~\ref{fig:MLP2}; the first will then act as a feature extractor and the newly added layer will then process only the limited number of features generated by the first. This alteration allows for the reduction in dimension of the first hidden layer, which reduces the total number of connections to the input layer. However, this also increases the network depth and results in a cost penalty (in terms of requiring additional memory accesses and arithmetic operations associated with the newly added layer). The key problem  demonstrated is that even for these two simple configurations, there is no systematic way to determine which design yields superior performance without having trained and evaluated both. Even when the designer has \textit{a priori} knowledge of the application, determining the optimal hyper-parameters is non-intuitive, especially for deep networks.

A concrete example of the described problem would be the design of an embedded system to recognize handwritten numerical characters (such as those contained in the \acs{MNIST} dataset). If the implementation goal is throughput of categorized digits, and a penalty is incurred when a character is misclassified (perhaps requiring manual intervention), then a smaller network that requires fewer clock cycles to evaluate may still result in an overall throughput exceeding those of more accurate alternatives. In such a scenario, the engineer would require knowledge of the cost and performance of all Pareto-optimal solutions in order to meet all requirements with the lowest implementation costs. 

\begin{figure*}
	\centering
	\includegraphics[width=0.63\linewidth]{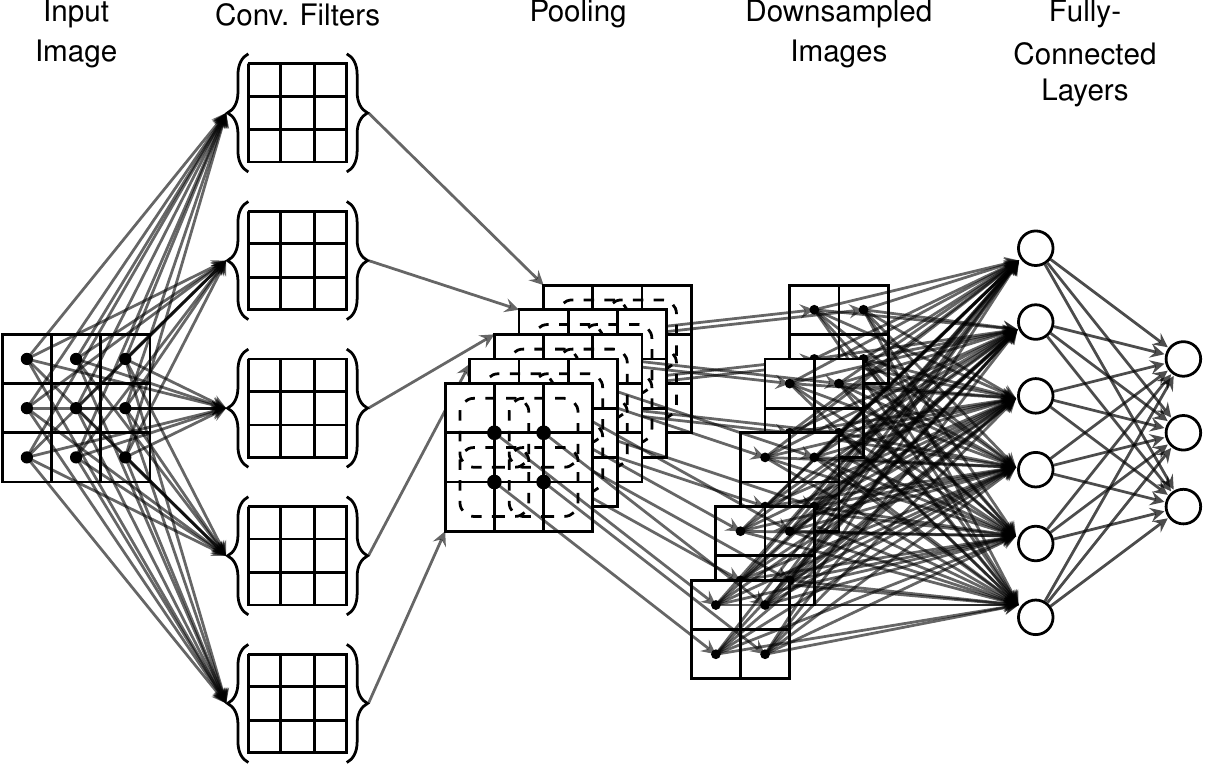}
	\caption{\acs{CNN} computational stages}
	\label{fig:CNN_graph}
\end{figure*}

Even more complex structures are those of \ac{CNN}, which have demonstrated state-of-the-art results in image recognition \cite{Krizhevsky:AlexNet}. The use of convolutional layers further complicates the design process as they introduce a separation between the memory and processing requirements. Illustrated in Figure~\ref{fig:CNN_graph}, convolutional layers are composed of trainable filters (five \textit{3-by-3} kernels are shown). \acp{CNN} have the advantage of reduced memory requirements due to each convolutional filter reusing the same kernel weights for all input values. However, this is at the expense of increased processing demands, the result of each convolutional filter requiring $N^2$ multiplication operations for each input value (where the convolutional kernels are sized \textit{N-by-N}). In a \ac{CNN}, each filter produces a processed copy of the input data (as illustrated in Figure~\ref{fig:CNN_graph}); and without any form of down-sampling, this greatly increases the computational complexity of the following layers. An example of a possible efficient down-sampling method is the inclusion of max-pooling (or mean-pooling) after convolutional layers \cite{Krizhevsky:AlexNet}. Illustrated in Figure~\ref{fig:CNN_graph}, max-pooling refers to partitioning the filtered images into non-overlapping \textit{K-by-K} regions, with each outputting the maximum pixel value within (alternatively, mean-pooling would output the mean value). All of these additional \ac{CNN} parameters further increase the design space dimensionality, forcing a designer to not only choose how many filters to use in each layer, but also the kernel and pooling sizes. This greatly affects both the performance and computational complexity of the resulting networks.

\subsection{Summary of Contributions}
\label{sec:contributions}

The key contributions of this work are presented as follows:
\begin{itemize}[topsep=0pt, partopsep=0pt]
	\item We introduce a \ac{DSE} method, which employs \ac{RSM} techniques to predict classification accuracy, to automate the design of \ac{ANN} hyper-parameters. This method is then validated with \ac{MLP} and \ac{CNN} designs targeting the \acs{CIFAR}-10 and \acs{MNIST} image recognition datasets \cite{Krizhevsky:CIFAR, LeCun:MNIST}.
	\item We demonstrate that multi-objective hyper-parameter optimization can successfully be used as a method to reduce \ac{ANN} implementation cost (computational complexity).
\end{itemize}

\fancyhf{}
\section{Related Work}
\label{sec:related_work}

This work exists at the intersection of two fields: automated hyper-parameter optimization, and reduction of \ac{ANN} computational complexity. To our knowledge, this work is the first that has applied automated hyper-parameter optimization as a method of reducing \ac{ANN} computational complexity.

\subsection{Hyper-Parameter Optimization}
\label{:sec:hyperparameter_optimization}

The design of neural network hyper-parameters has long been considered to be unwieldy, unintuitive, and as a consequence, ideal for automated hyper-parameter optimization techniques such as in \cite{Bergstra:HyperRandomSearch}, \cite{Bergstra:HyperAlgorithms}, \cite{Domhan:HyperLearningCurves}, and \cite{Young:HyperEvolutionary}. However, these works have been developed with the sole purpose of optimizing performance, with little regard to the resulting computational resource requirements.

Two types of sequential hyper-parameter optimization algorithms were presented in \cite{Bergstra:HyperAlgorithms}; in both cases experimental results compared positively to human designed alternatives. These findings were echoed in \cite{Bergstra:HyperRandomSearch}, where it was demonstrated that a random search of a large design space, which contains areas that may be considered less promising, can at times exceed the performance of manually tuned hyper-parameters.

Positive results were also presented in \cite{Domhan:HyperLearningCurves} where similar algorithms were extended using a method to extrapolate the shape of learning curves during training, so as to evaluate fewer epochs for unfit solutions and reduce design time. Since parameters which perform favourably for a large network may not be optimal for a smaller alternative, and the most important hyper-parameters (with the greatest impact on resulting performance) may vary for different data sets, the need for multi-objective optimization (especially where low power platforms are targeted) is clear \cite{Bergstra:HyperRandomSearch, Young:HyperEvolutionary}.

Additionally, in \cite{Palermo:ReSPIR} an automated \ac{DSE} method was applied to the multi-objective optimization problem of application-specific \acs{MPSoC} design, a field which also consists of high-dimensionality solution spaces. It was demonstrated that through \ac{RSM} the presented \ac{DSE} method could efficiently identify Pareto-optimal architectures.

\subsection{Weight Quantization and Pruning}
\label{sec:quantization_pruning}

There also exists a body of work attempting to reduce the computational complexity of \ac{ANN} models through weight quantization or the removal of extraneous node connections (pruning). The research in \cite{Courbariaux:BinaryConnect} and \cite{Dally:Cost} are examples of two methods that construct networks which reduce the need for multiplications, or modify already trained networks in order to minimize the number of non-zero weights. 

In \cite{Courbariaux:BinaryConnect}, the authors attempted to reduce the computational resources required when evaluating fully-connected, as well as convolutional, networks through representing all weights as binary values of $+1$ or $-1$. Doing so reduces the number of multiplication operations performed each forward pass; however the requirement to store floating-point weights during training remains. The work in \cite{Courbariaux:BinaryConnect} also compared trained binary weighted networks to traditional equivalents with equal layer dimensions and similar performance was demonstrated. However, \cite{Courbariaux:BinaryConnect} only considered very large network dimensions;  further benefits may be obtained from smaller optimized architectures.

Instead of restricting weights to specific values, in \cite{Dally:Cost} a pruning method was presented in which a network can be trained while reducing the number of non-zero weights. The resulting compressed networks have lower bandwidth requirements and require fewer multiplications due to most weights being zero. The results in \cite{Dally:Cost} demonstrated up to 70\% reductions in the numbers of floating-point operations required for various networks, with little to no reduction in performance. However, such pruning methods are not mutually exclusive to the use of \ac{DSE} tools and could very well be implemented in conjunction with the presented methodology in order to compress an already optimized network configuration.

\section{\acs{ANN} Self-Design Methodology}
\label{sec:DSE}

This work presents a \ac{DSE} approach that searches for Pareto-optimal hyper-parameter configurations and has been applied to both \ac{MLP} and \ac{CNN} topologies. The design space is confined to: the numbers of \ac{FC} and convolutional layers, the number of nodes or filters in each layer, the convolution kernel sizes, the max-pooling sizes, the type of activation function, and network training rate. These degrees of freedom constitute vast design spaces and all strongly influence the resulting networks' cost and performance.

For design spaces of such size, performing an exhaustive search is intractable (designs with over $10^{10}$ to $10^{20}$ possible solutions are not uncommon), therefore we propose to model the response surface using an \ac{ANN} for regression where the set of explored solution points is used as a training set. The presented meta-heuristic \ac{ANN} is then used to predict the performance of candidate networks; only points which are expected not to be Pareto-dominated are explored.

\subsection{Main \acs{DSE} Algorithm Overview}
\label{sec:DSE_algorithm}

\begin{figure}
	\centering
	\includegraphics[width=0.75\columnwidth]{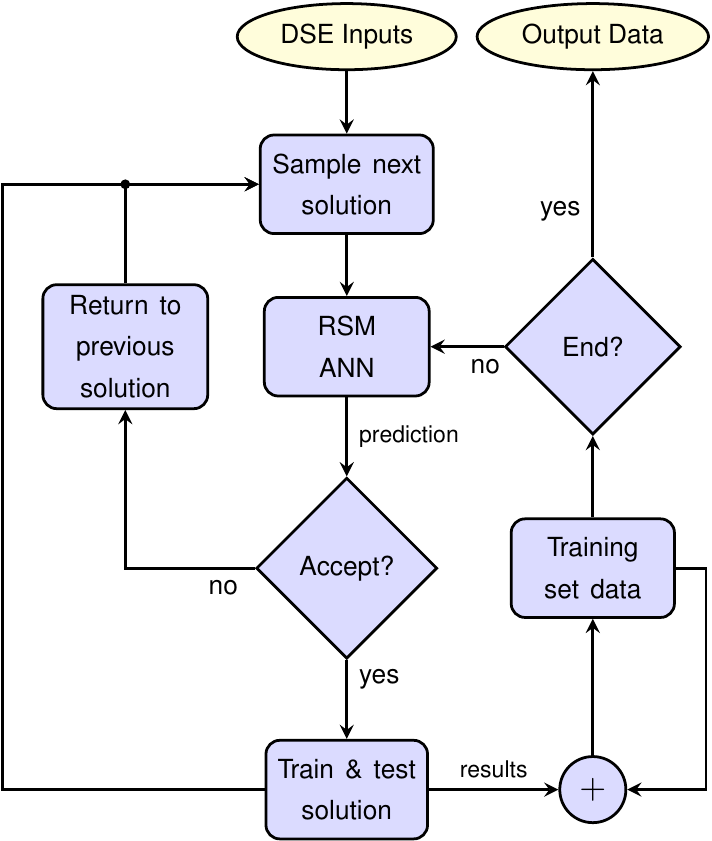}
	\caption{\acs{DSE} algorithm flow}
	\label{alg:DSE}
\end{figure}

A flowchart describing the \ac{DSE} implementation is shown in Figure~\ref{alg:DSE}. The general steps performed during the design space exploration can be broken down into:

\begin{enumerate}[topsep=0pt, partopsep=0pt, parsep=0pt]
	\item Sample the next candidate point from a Gaussian distribution centred around the previously explored solution (or sample a random point for the first iteration).
	\item Predict the candidate solution performance using the \ac{RSM} neural network, and calculate the cost as: \vspace{-1.5mm}
	\begin{center}
		$cost = \left( \text{\# of weights} \right) \times \left( \text{weight unit cost} \right) + \left( \text{\# of multiplications} \right) \times \left( \text{multiplication unit cost} \right)$
	\end{center}
	\item Compare the predicted results to the current Pareto-optimal front. If the candidate is predicted to be Pareto-dominated, it is accepted with probability $\alpha$, otherwise it is accepted with probability $1-\alpha$.
	\item If rejected, the previously explored solution is rolled back and the algorithm returns to Step 1.
	\item If accepted, the candidate model is trained, tested, and the evaluated results are added to the training set of the \ac{RSM} \ac{ANN} (which is then retrained).
	\item Finally, if the training set size exceeds the maximum number of desired iterations, the process ends. Otherwise, the algorithm returns to Step 1 and a new solution is sampled.
\end{enumerate}

\subsection{Candidate Solution Sampling}
\label{sec:sampling_method}

The sampling strategy proposed is an adaptation of the Metropolis-Hastings algorithm \cite{Metropolis–Hastings}. In each iteration a new candidate is sampled from a Gaussian distribution centred around the previously explored solution point. Performing this random walk limits the number of samples chosen from areas of the design space that are known to contain unfit solutions, thereby reducing wasted exploration effort. However, exploring an inferior solution may eventually lead to those of superior performance, therefore the probability of accepting such a solution ($\alpha$) must remain greater than zero; this also ensures that the training set for the \ac{RSM} \ac{ANN} remains varied. All experimental results in Section~\ref{sec:experimental_results} were obtained with $\alpha = 10^{-4}$.

\subsection{Predictive Neural Network Design}
\label{sec:RSM}

We choose to model the response surface using a \ac{MLP} model with an input set representative of network hyper-parameters and a single output trained to predict the error of corresponding networks. This \ac{RSM} \ac{ANN} is composed of two hidden \ac{ReLU} layers and a linear output layer. Experimental results demonstrated that sizing the hidden layers with $25\times$ to $30\times$ the number of input nodes provided the best performance.

The \ac{RSM} network inputs are formed as arrays characterizing all explored dimensions. Integer input parameters (such as number of nodes in a hidden layer, or size of the convolutional kernels) are scaled by the maximum possible value of the respective parameter, resulting in normalized variables between 0 and 1. For each parameter that represents a choice where the options have no numerical relation to each other (such as whether \ac{ReLU} or sigmoid functions are used) an input mode is added and the node that represents the chosen option is given an input value of 1, all others $-1$. For example, a solution with two hidden layers with 20 nodes each (assuming a maximum of 100), using \acp{ReLU} (with the other option being sigmoid functions) and with a learning rate of 0.5 would be presented as input values: $\left[ 0.2, 0.2, 1, -1, 0.5 \right]$.

The \ac{RSM} model was trained using \ac{SGD}, where 100 training epochs were performed on the set of explored solutions each time the next is evaluated (and in turn, added to the training set). The learning rate was kept constant, with a value of 0.1, in order to train the network quickly during early exploration, when the set of evaluated solutions is limited.

\section{Experimental Setup}
\label{sec:experimental_setup}

We evaluated our \ac{DSE} strategy on \ac{ANN} applications targeting the standard \acs{MNIST} and \acs{CIFAR}-10 image recognition datasets. In the case of designing \ac{MLP} models targeting the simpler \acs{MNIST} problem, the results generated by the \ac{DSE} algorithm were compared to the true Pareto-optimal front obtained from an exhaustive search performed on a constrained solution space. Such a limitation of the design space was required in order to render the exhaustive search tractable. The  \ac{DSE} algorithm was also evaluated on much larger design spaces for both \ac{MLP} and \ac{CNN} models targeting \acs{MNIST} as well as an even larger one for the design of \acp{CNN} targeting \acs{CIFAR}-10. In all cases only a few iterations were required in order to converge to approximated Pareto-optimal fronts.

In order to perform the \ac{DSE}, all \ac{ANN} models were trained and tested using the \textit{Theano} framework in order to take advantage of \acs{GPGPU} optimizations \cite{Bengio:Theano}. This allowed the entire \ac{DSE} and evaluation to be performed using a Nvidia \textit{Tesla K20c} \acs{GPU}. Upon completion, all explored network data and simulation results were stored to disk, allowing future design re-use.

To model cost, we assumed normalized memory access and multiply-accumulate operation costs. These can be quantified at either the software or hardware levels (depending on the target application). In both cases the costs (such as power, area, or time) are implementation specific and the exact values used are transparent to the \ac{DSE} algorithm. Normalized costs assumed for all the experimental results are shown in Table~\ref{table:costs}, which are based on the energy costs for 32-bit floating-point operations from \cite{Dally:Cost}.

\begin{table}
	\caption{Normalized cost values}
	\centering
	\begin{tabular}{m{0.275\columnwidth}<{\centering} m{0.275\columnwidth}<{\centering} m{0.275\columnwidth}<{\centering}}
		\toprule
		Operation & Energy Cost from \cite{Dally:Cost} & Normalized Cost \\
		\toprule
		Addition & $0.9\text{pJ}$ & N/A \\
		\midrule
		Multiplication & $3.7\text{pJ}$ & N/A \\
		\midrule
		Multiply-Accumulate & $0.9\text{pJ} + 3.7\text{pJ}$ & $1$ \\
		\midrule
		DRAM Memory Access & $640\text{pJ}$ & $139$ \\
		\bottomrule
	\end{tabular}
	\label{table:costs}
\end{table}

We experimentally validated the \ac{DSE} algorithm on two separate image recognition databases; future work is planned to address a broader range of benchmarks. The first, \acs{MNIST}, contains $28\times28$ pixel grayscale images of handwritten numeric characters (from 0 to 9). The second, \acs{CIFAR}-10, is a much more complicated dataset composed of $32\times32$ RGB colour images, each belonging to one of ten object categories.

\section{Results}
\label{sec:experimental_results}

The network hyper-parameters composing the design spaces explored are outlined in Table~\ref{table:DSE_parameters}. In addition, several parameters were kept constant for all experiments: output layers were composed of \textit{softmax} units (with \textit{jth} output defined as ${e^{{in}_j}}\div{\sum_{k=1}^{K} e^{{in}_k}}$ for a layer with $K$ nodes) and all network training was performed with categorical cross-entropy $\left(-\sum \left[ targets \times log \left( predictions \right) \right]\right)$ as loss function \cite{Hinton:softmax, Jarrett:BestMultiStageArchitecture}. Finally, for all except the reduced \acs{MNIST} design problem, batch normalization was included after each network layer in order to smooth the response surfaces \cite{Ioffe:BatchNorm}.

\begin{table}[t]
	\caption{Experimental design-space parameters}
	\centering
	\resizebox{\columnwidth}{!}{
		\begin{tabular}{ccccc}
			\toprule
			Target & Parameter & Range & Steps & Scale \\
			\toprule
			\multirow{7}{1.45cm}{\centering Restricted \acs{MNIST} (\acs{MLP})} & Number of \acs{FC} layers & 1 to 3 & 3 & Linear \\
			& Nodes per \acs{FC} layer & 10 to 200 & 10 & Log \\
			& Learning rate & 0.01 to 0.8 & 9 & Log \\
			& Activation function & \acs{ReLU} & 1 & N/A \\
			& Training algorithm & \acs{SGD} & 1 & N/A \\
			& Batch sizes & 200 & 1 & N/A \\
			& Training epochs & 10 & 1 & N/A \\
			\midrule
			\multirow{7}{1.45cm}{\centering \acs{MNIST} (\acs{MLP})} & Number of \acs{FC} layers & 1 to 3 & 3 & Linear \\
			& Nodes per \acs{FC} layer & 10 to 500 & 85 & Log \\
			& Learning rate & 0.001 to 0.8 & 16 & Log \\
			& Activation function & \acs{ReLU} or tanh & 2 & N/A \\
			& Training algorithm & \acs{SGD} & 1 & N/A \\
			& Batch sizes & 200 & 1 & N/A \\
			& Training epochs & 10 & 1 & N/A \\
			\midrule
			\multirow{11}{1.45cm}{\centering \acs{MNIST} (\acs{CNN})} & Number of \acs{CNN} layers & 1 to 2 & 2 & Linear \\
			& Number of \acs{FC} layers & 1 to 2 & 2 & Linear \\
			& Filters per \acs{CNN} layer & 8 to 128 & 8 & Log \\
			& Nodes per \acs{FC} layer & 10 to 250 & 16 & Log \\
			& Learning rate & 0.01 to 0.8 & 12 & Log \\
			& Filter kernel size & 1x1 to 5x5 & 3 & Linear \\
			& Max-pool size & 2x2 to 4x4 & 3 & Linear \\
			& Activation function & \acs{ReLU} & 1 & N/A \\
			& Training algorithm & \acs{SGD} & 1 & N/A \\
			& Batch sizes & 200 & 1 & N/A \\
			& Training epochs & 10 & 1 & N/A \\
			\midrule
			\multirow{11}{1.45cm}{\centering \acs{CIFAR}-10 (\acs{CNN})} & Number of \acs{CNN} layers & 1 to 3 & 3 & Linear \\
			& Number of \acs{FC} layers & 1 to 2 & 2 & Linear \\
			& Filters per \acs{CNN} layer & 16 to 512 & 20 & Log \\
			& Nodes per \acs{FC} layer & 10 to 500 & 16 & Log \\
			& Learning rate & 0.001 to 0.1 & 16 & Log \\
			& Filter kernel size & 3x3 to 9x9 & 4 & Linear \\
			& Max-pool size & 2x2 to 3x3 & 2 & Linear \\
			& Activation function & \acs{ReLU} & 1 & N/A \\
			& Training algorithm & \textit{Adam} \cite{Kingma:Adam} & 1 & N/A \\
			& Batch sizes & 200 & 1 & N/A \\
			& Training epochs & 20 & 1 & N/A \\
			\bottomrule
		\end{tabular}
	}
	\label{table:DSE_parameters}
\end{table}

\subsection{Exhaustive Search Comparison}
\label{sec:exhaustive}

In order to evaluate the efficiency with which the method approximates the true Pareto-optimal front, we first compare experimental results to those of an exhaustive search targeting the design of \ac{MLP} models for the \acs{MNIST} dataset. In order to make an exhaustive search tractable, we limited the design space to only the values outlined in the first section of Table~\ref{table:DSE_parameters}. This resulted in a moderate design space of $10^4$ solutions, all of which were trained and tested.

\begin{figure}
	\centering
	\includegraphics[width=\columnwidth]{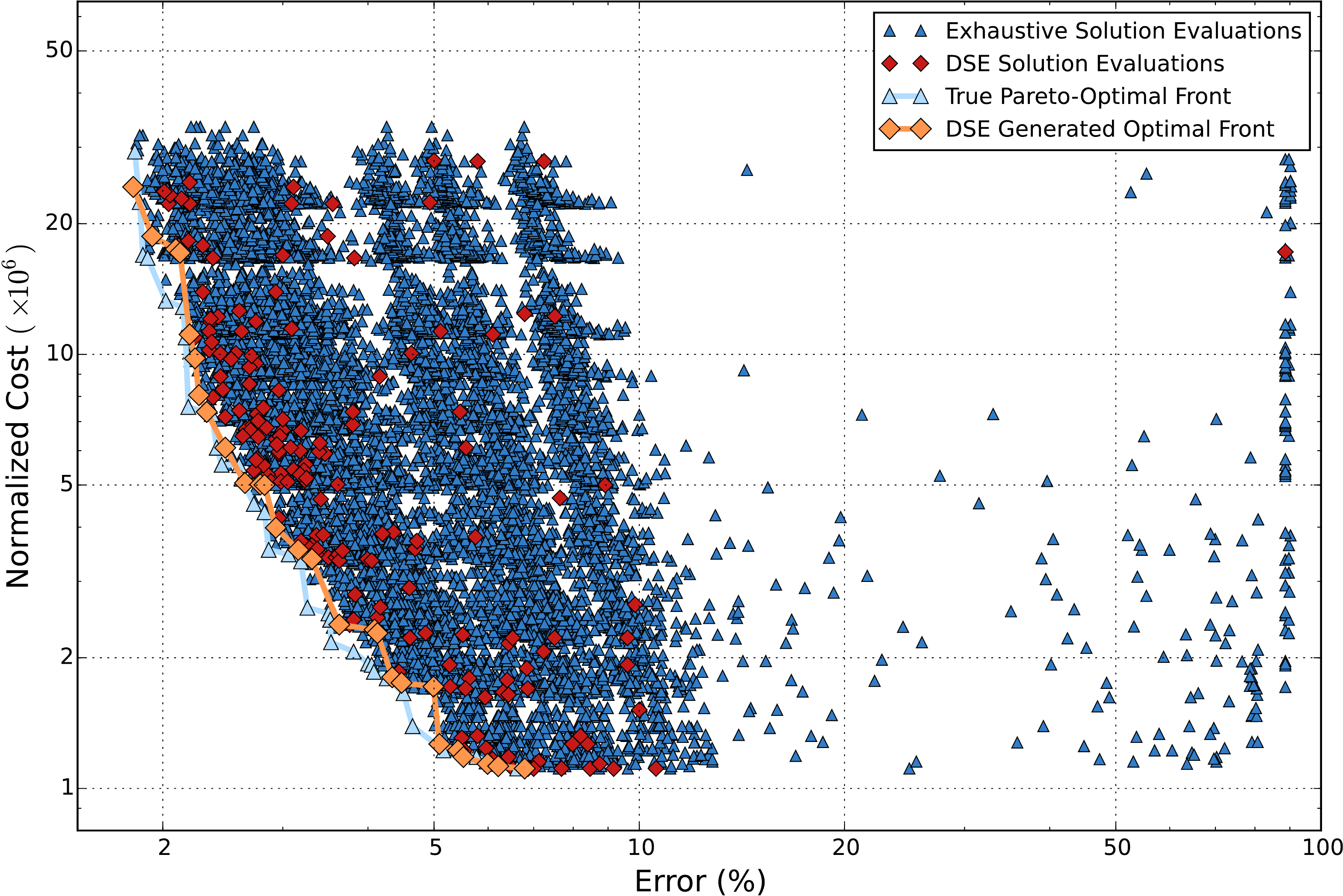}
	\caption{Exhaustive search versus \acs{DSE} results}
	\label{fig:MNIST_exhaustive}
\end{figure}

The results of executing the \ac{DSE} algorithm for 200 iterations (each iteration represents a design training, evaluation, and model update pass) are plotted alongside those of the exhaustive search in Figure~\ref{fig:MNIST_exhaustive}. These results demonstrate that the true Pareto-optimal front is very closely approximated by the presented method, while requiring very few solutions to be evaluated. However, it should be noted that \ac{SGD} is by nature non-deterministic and training the same network twice may yield different performances. Consequently, the \ac{DSE} generated results (shown in Figure~\ref{fig:MNIST_exhaustive}) dominate those of the exhaustive search at several points. This figure also demonstrates that the majority of solutions evaluated by the \ac{DSE} algorithm remain within a close vicinity of the true Pareto-front, successfully avoiding Pareto-dominated areas. Since the true Pareto-optimal front is known for this restricted case, we use \ac{ADRS} as the metric of evaluation; quantifying how closely the approximated set differs from the exact \cite{Palermo:ReSPIR}. \ac{ADRS} is defined in Eq.~\eqref{eqn:ADRS}, where $\Lambda$ and $\Pi$ are the approximate and exact Pareto-optimal fronts, and the $\delta \left( x_R, x_A \right)$ function represents the normalized distance between solutions $x_R$ and $x_A$.

\begin{eqnarray}
	\label{eqn:ADRS}
	ADRS(\Pi, \Lambda) = \frac{1}{|\Pi|}\sum_{x_R \in \Pi} \min_{x_A \in \Lambda} \Big( \delta \left( x_R, x_A \right) \Big)
\end{eqnarray}

The evolution of the approximated Pareto-optimal set discovered by the \ac{DSE} algorithm, as a function of iterations completed is plotted in Figure~\ref{fig:MNIST_MLP_reduced_results}. Evident from this figure is that the optimal front obtained from the \ac{DSE} algorithm progressively approaches the true Pareto-optimal, while evaluating only a comparatively small number of iterations. The proposed method identifies the optimal solutions with high accuracy, achieving low \ac{ADRS} values of 7.1\% after completing only 50 iterations, 5.0\% after 100, and 3.6\% after 200. The results in Table~\ref{table:ADRS} also demonstrate fast convergence, with the largest changes occurring over the first 30 iterations. Further execution yields more gradual changes as the approximated Pareto-front asymptotically approaches the exact. Decreasing design time (both in terms of computation and man-hours), these results also demonstrate that by exploring less than 1\% of the design space, the \ac{DSE} method closely predict which hyper-parameters are required for optimal solutions. In addition, the generated set of Pareto-optimal configurations also exposes the trade-offs application designers may want to make for cost-constrained systems, allowing for more informed design choices to be made.

\begin{table}
	\caption{Evolution of \acs{ADRS} values}
	\centering
	\begin{tabular}{m{0.2\columnwidth}<{\centering} m{0.225\columnwidth}<{\centering} m{0.2\columnwidth}<{\centering}}
		\toprule
		\acs{DSE} Iterations & \acs{ADRS} of Explored Set & Execution Time \\
		\toprule
		10 & 30\% & 2.0 min \\
		\midrule
		20 & 21\% & 4.1 min \\
		\midrule
		30 & 8.4\% & 6.4 min \\
		\midrule
		50 & 7.1\% & 11.2 min \\
		\midrule
		70 & 6.1\% & 16.4 min \\
		\midrule
		100 & 5.0\% & 25.4 min \\
		\midrule
		150 & 4.5\% & 45.2 min \\
		\midrule
		200 & 3.6\% & 70.1 min \\
		\bottomrule
	\end{tabular}
	\label{table:ADRS}
\end{table}

\subsection{Evaluation on Expanded Design Spaces}
\label{sec:results_expanded}

\begin{figure}
	\centering
	\subfigure[Pareto-optimal results (restricted \acs{MNIST} \acs{MLP} design-space)]{
		\includegraphics[width=0.85\columnwidth]{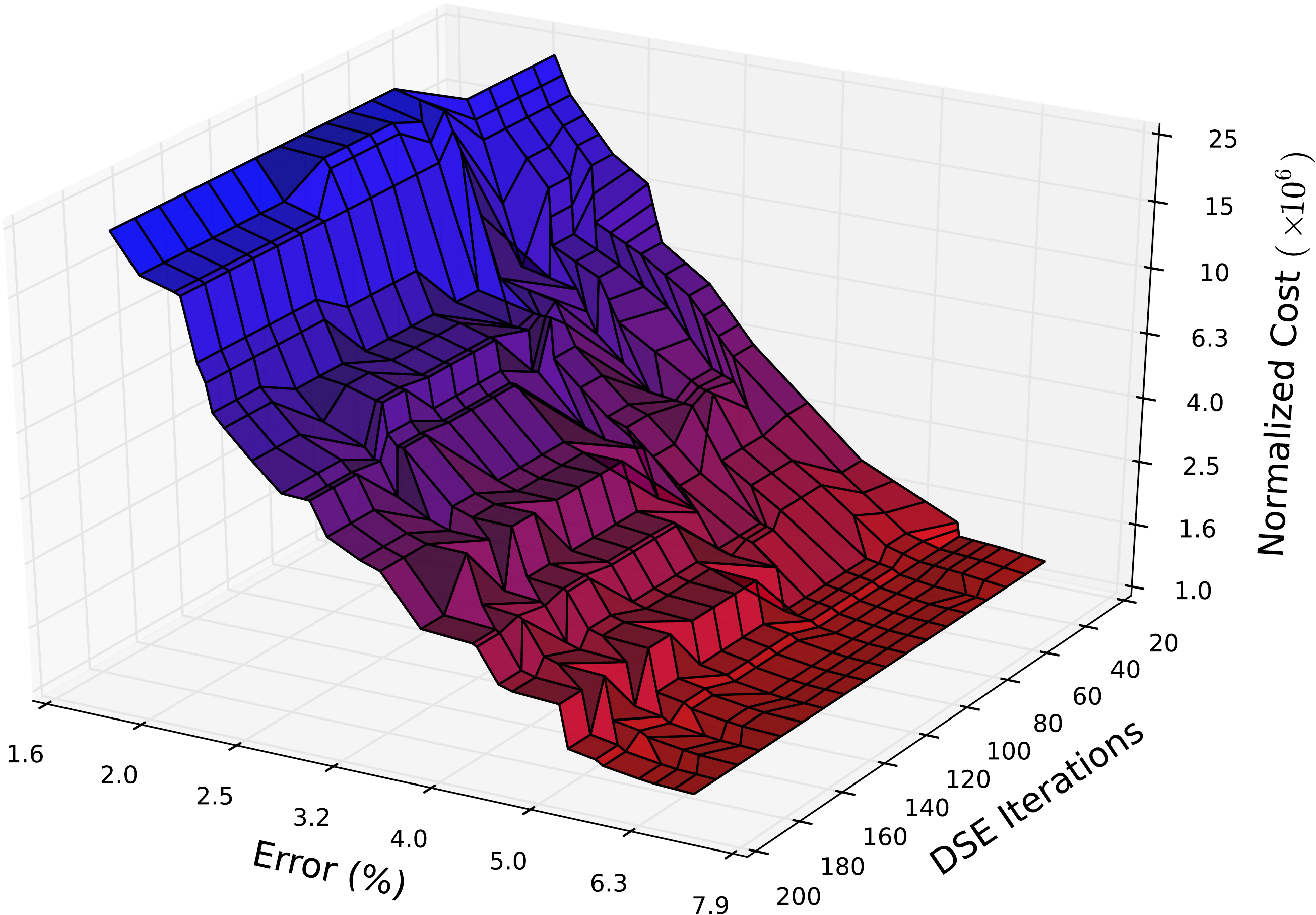}
		\label{fig:MNIST_MLP_reduced_results}
	}
	\subfigure[Pareto-optimal results (\acs{MNIST} \acs{MLP})]{
		\includegraphics[width=0.85\columnwidth]{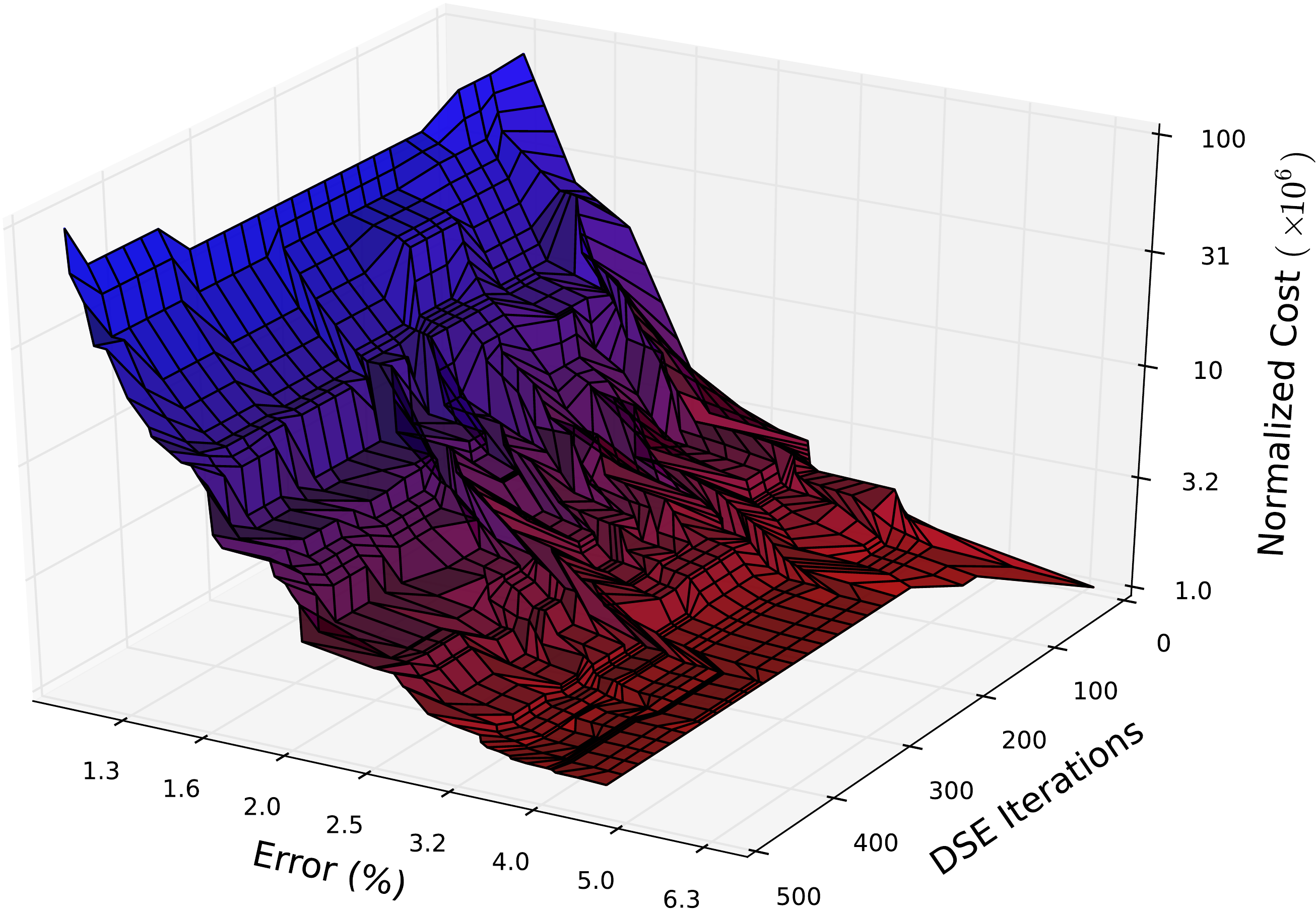}
		\label{fig:MNIST_MLP_results}
	}
	\subfigure[Pareto-optimal results (\acs{MNIST} \acs{CNN})]{
		\includegraphics[width=0.85\columnwidth]{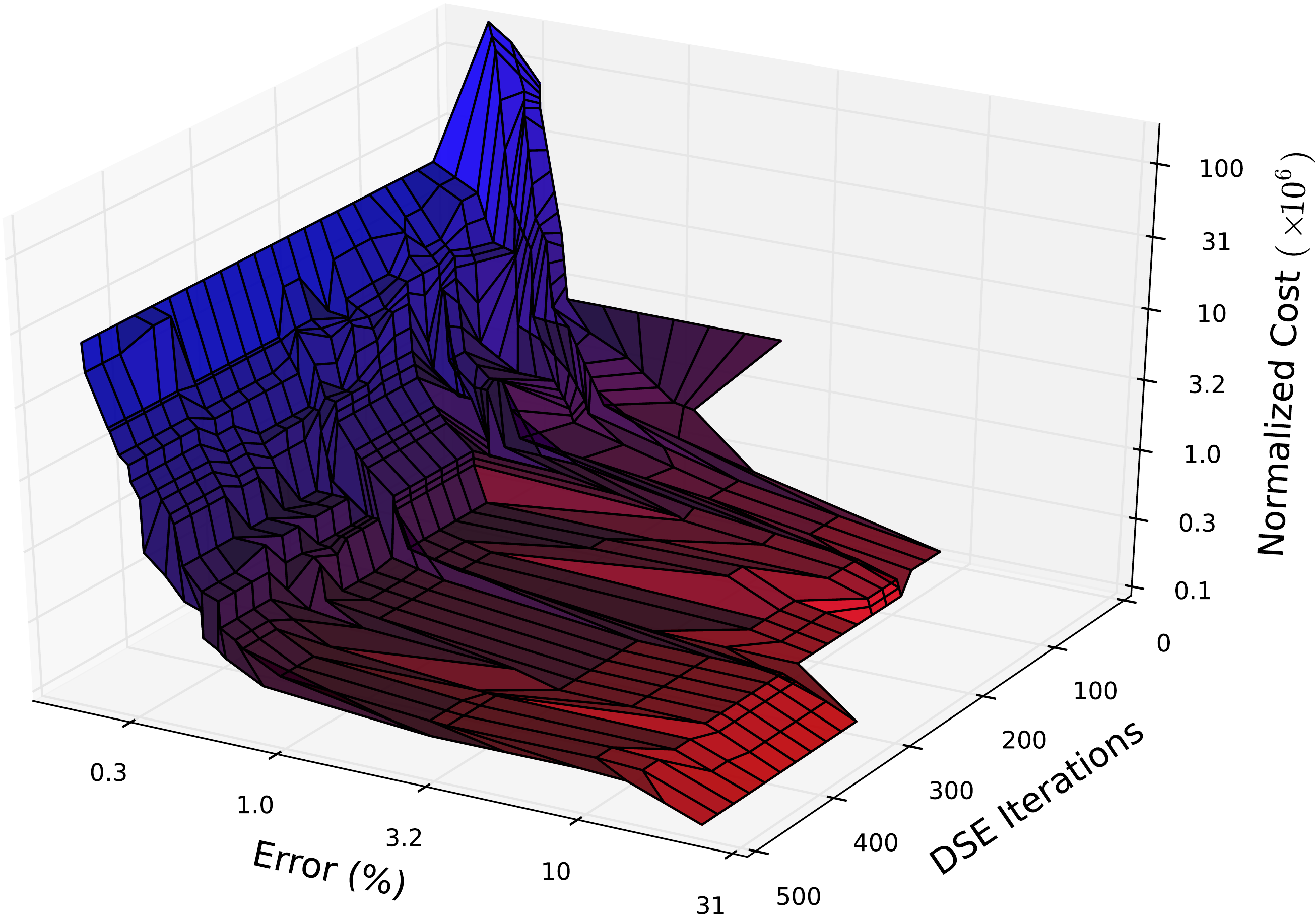}
		\label{fig:MNIST_CNN_results}
	}
	\subfigure[Pareto-optimal results (\acs{CIFAR}-10 \acs{CNN})]{
		\includegraphics[width=0.85\columnwidth]{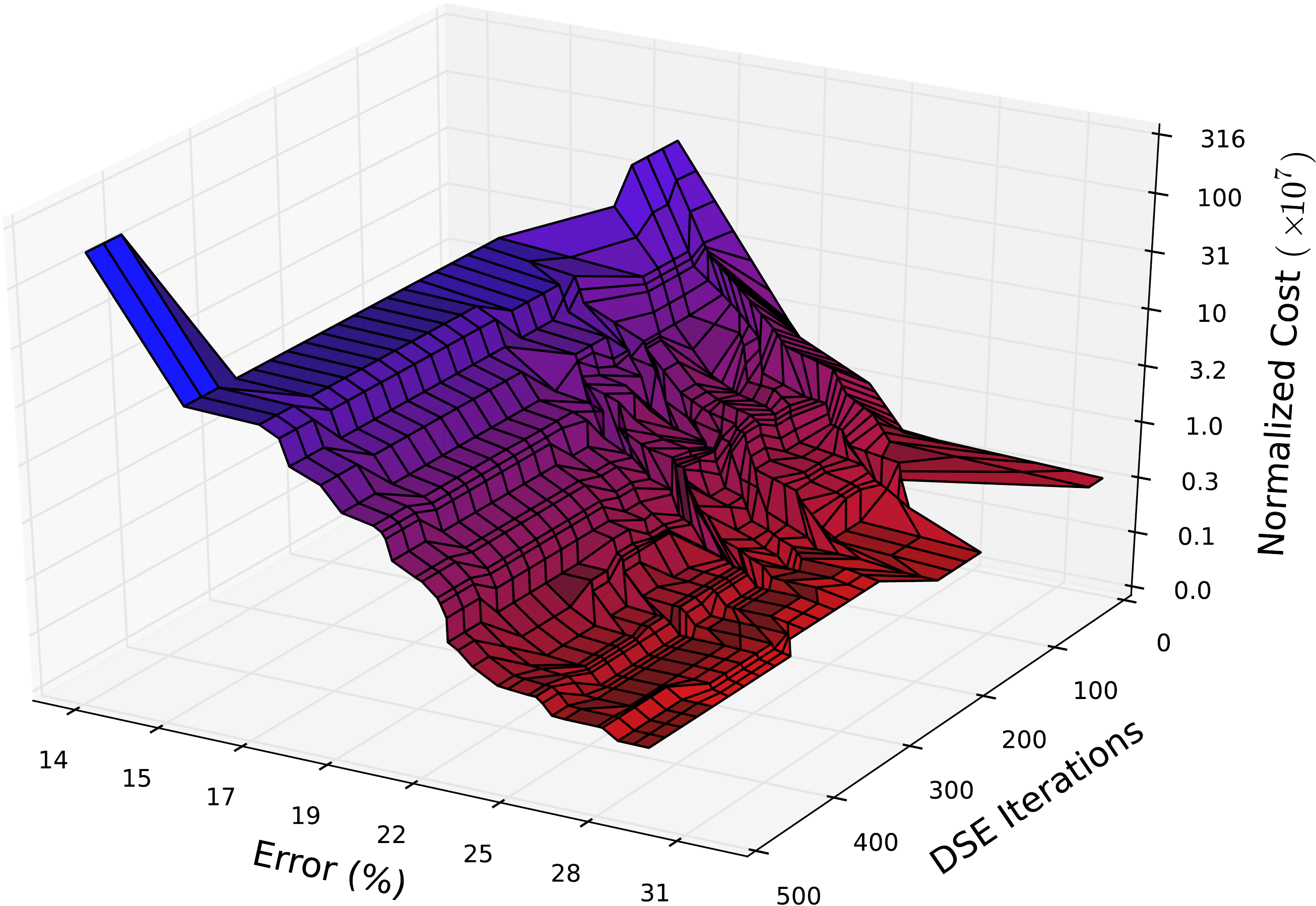}
		\label{fig:CIFAR_CNN_results}
	}
	\caption{\acs{DSE} generated experimental results}
	\label{fig:experimental_results}
\end{figure}

In order to evaluate performance of the heuristic method for much larger designs, the algorithm was run on the remaining spaces described in Table~\ref{table:DSE_parameters} for both \ac{MLP} and \ac{CNN} design problems. The total execution times (on an Intel \textit{Xeon E5-1603} \acs{CPU} with 8GB of \acs{RAM} and a Nvidia \textit{Tesla K20c} \acs{GPU}) for the design examples are listed in Table~\ref{table:runtime} and the corresponding Pareto-optimal results (plotted as functions of the number of iterations completed) are shown in Figure~\ref{fig:experimental_results}. In these plots, the colour scheme presents solutions with low error in blue, and low cost in red. Even though the \ac{DSE} outputs cannot be directly compared to the true results from an exhaustive search (the \acs{CIFAR}-10 example design space exceeds $10^{10}$ solutions), the trends discussed in Section~\ref{sec:exhaustive} are mirrored by all plots in Figure~\ref{fig:experimental_results}.

\begin{table}
	\caption{\acs{DSE} execution times}
	\centering
	\begin{tabular}{m{0.27\columnwidth}<{\centering} m{0.29\columnwidth}<{\centering} m{0.29\columnwidth}<{\centering}}
		\toprule
		Target & \acs{DSE} Algorithm Execution Time & Mean Solution Evaluation Time \\
		\toprule
		\acs{MNIST} (\acs{MLP}) & 3.6 h & 30 s \\
		\midrule
		\acs{MNIST} (\acs{CNN}) & 18 h & 2 min \\
		\midrule
		\acs{CIFAR}-10 (\acs{CNN}) & 66 h & 8 min \\
		\bottomrule
	\end{tabular}
	\label{table:runtime}
\end{table}

Because of the intractable nature of such exhaustive searches, the \ac{DSE} generated results are not expected to always predict the true Pareto-optimal fronts. Instead, automated exploration must only match, or exceed, the fitness of manually designed networks in order to provide \ac{NRE} cost reductions. In comparison with manually designed architectures in literature, the Pareto-optimal results in Figure~\ref{fig:MNIST_CNN_results} include points that offer equivalent performance to the \ac{CNN} designs in \cite{Jarrett:BestMultiStageArchitecture} and \cite{LeCun:MNIST}, with implementation costs as low as 25\% of their manually designed counterparts (when weighted with the same cost model detailed in Table~\ref{table:costs}). This demonstrates that the proposed \ac{DSE} method can design \acp{ANN} with vastly reduced cost requirements and same levels of performance when compared to manual design approaches.

Furthermore, the Pareto-optimal results can also find data points with substantial cost savings penalized only by a small decrease in performance. Given that increasing accuracy by only 0.01\% requires a doubling in implementation cost (for \acs{MNIST} \ac{CNN} designs shown in Figure~\ref{fig:MNIST_CNN_results}), these additional points are invaluable for application platforms with extremely stringent cost budgets.

\subsection{\acs{RSM} Prediction Accuracy}
\label{sec:RSM_accuracy}

\begin{figure}[h]
	\centering
	\includegraphics[width=\columnwidth]{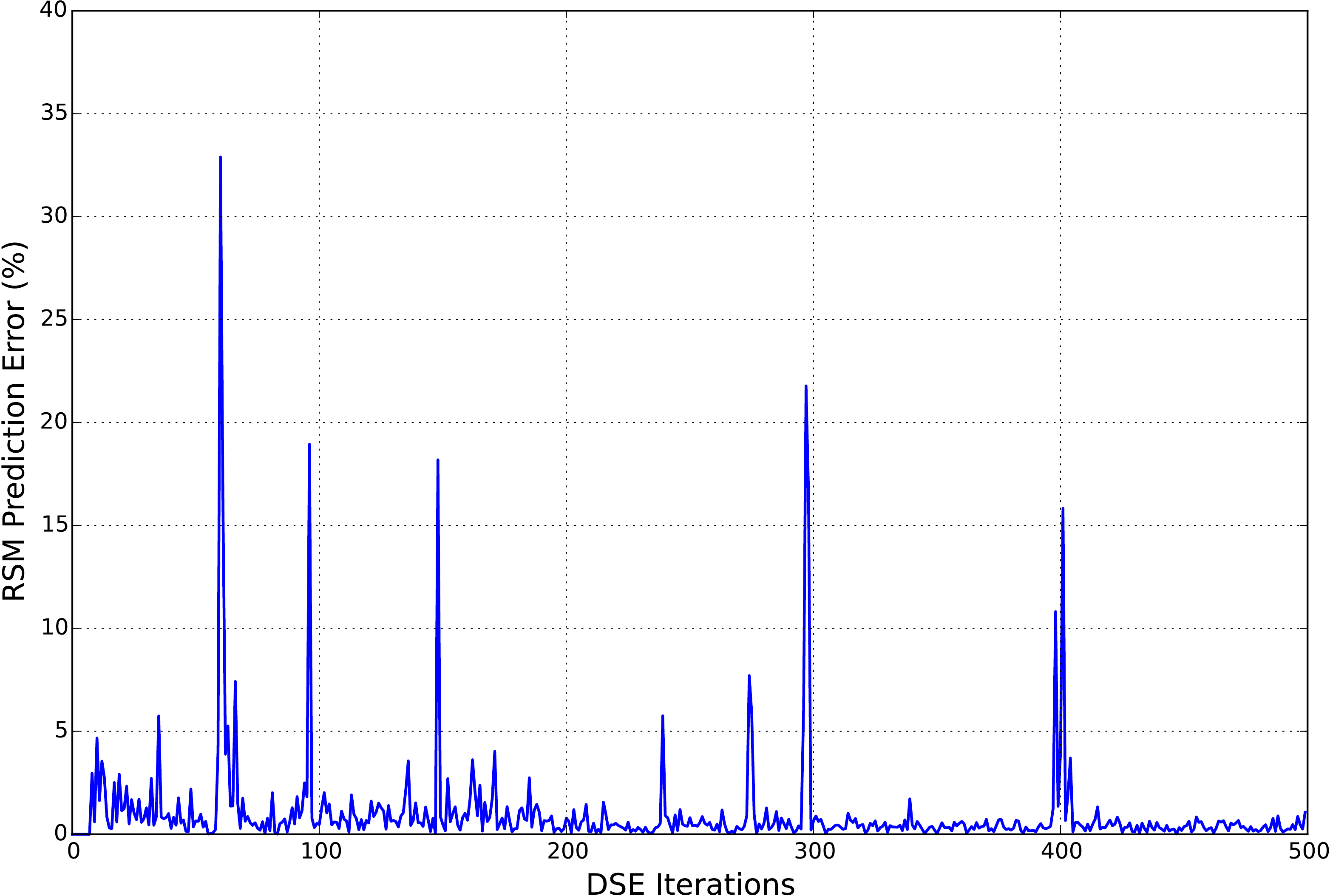}
	\caption{\acs{RSM} error during \acs{MNIST} \acs{CNN} design}
	\label{fig:RSM_error}
\end{figure}

In order to validate the assumption that a neural network can be trained through regression to model the response surface with sufficient accuracy, the \ac{RSM} \ac{ANN} prediction error is plotted in Figure~\ref{fig:RSM_error}. This graph plots the absolute value of the error (\% difference between the predicted and the evaluated performance of each explored solution) for each of the 500 \ac{DSE} algorithm iterations performed during the \acs{MNIST} \ac{CNN} design example (with results in Figure~\ref{fig:MNIST_CNN_results}). The narrow error spikes, which are expected, occur at points where the \ac{DSE} algorithm encounters previously unexplored areas. As these solutions are added to the training set, the prediction error decreases as the response surface model is updated, and the spikes begin to occur less frequently. Outside of these sparse peaks, the prediction accuracy is exceptionally high; the mean error over the last 95 iterations (all points after the last spike) is only 0.35\%.

\section{Conclusion}
\label{sec:conclusion}

A \ac{DSE} method to automate the multi-objective optimization of neural network hyper-parameters, reducing both algorithm error and computational complexity, was presented. When compared to the results of an exhaustive search on a restricted design space targeting the \acs{MNIST} dataset, the presented method was shown to dramatically reduce computation time required for convergence to a Pareto-optimal solution set. A low \ac{ADRS} of 5\% was achieved after only 100 iterations; in practice, fewer solution evaluations may be required, with corresponding execution times less than those in Table~\ref{table:runtime}. Furthermore, scalability of the method was demonstrated on larger design spaces for both \ac{MLP} and \ac{CNN} models targeting the \acs{CIFAR}-10 dataset as well.

Even when evaluated on massive design spaces, the presented \ac{DSE} method was found to still efficiently converge to a diverse Pareto-optimal front. Not only was the automation of \ac{ANN} hyper-parameter design process demonstrated to be both feasible and time efficient, but when compared to manually designed networks from literature, the automated \ac{DSE} technique produced results with near identical performance while reducing the associated costs by a factor of $3\times$. As applications for \acp{ANN} make further inroads in mobile and embedded market segments, the need to reduce time-to-market and \ac{NRE} costs will necessitate the use of such automated design methods.

\section{Acknowledgements}
This work was supported in part by a \ac{PGS-D} scholarship from the \ac{NSERC}, as well as equipment donations from Nvidia Corporation.
\balance



\end{document}